\documentclass[letterpaper, 10 pt, conference]{ieeeconf}

\IEEEoverridecommandlockouts                              

\usepackage{amsmath}
\usepackage{amssymb}
\usepackage{balance}
\usepackage[noadjust]{cite}
\usepackage{color}
\usepackage{float}
\usepackage{gensymb}
\usepackage{graphicx}
\usepackage{mathtools}
\usepackage{multirow}
\usepackage{subcaption}
\usepackage{url}
\usepackage{xspace}
\usepackage{pifont}
\usepackage{placeins}
\usepackage{flushend}

\definecolor{samarth_color}{rgb}{.6,.4,.05}
\definecolor{ankur_color}{rgb}{.5,.7,.1}
\definecolor{andrew_color}{rgb}{0,0.35,0}
\definecolor{matthias_color}{rgb}{0,0,1}
\definecolor{vladlen_color}{rgb}{0,0,0.8}
\newcommand{\comment}[1]{} 

\newcommand{\T}[2]{^{#1}T_{#2}}

\newcommand{\citet}{\cite}
\newcommand{\citep}{\cite}

\usepackage{booktabs}
\makeatletter
\DeclareRobustCommand\onedot{\futurelet\@let@token\@onedot}
\def\@onedot{\ifx\@let@token.\else.\null\fi\xspace}

\def\eg{\emph{e.g}\onedot} 
\def\ie{\emph{i.e}\onedot}

\def\wrt{w.r.t\onedot} 

\makeatother

\newcommand{\YesV}{\ding{51}}%
\newcommand{\NoX}{\ding{55}}%
\newcommand\mypara[1]{\vspace{1mm}\noindent\textbf{#1}}

\makeatletter
\def\ignorecitefornumbering#1{%
     \begingroup
         \@fileswfalse
         #1
    \endgroup
}
\makeatother

\title{\LARGE \bf Zero-Shot Transfer of Haptics-Based Object Insertion Policies}

\author{Samarth Brahmbhatt$^{1}$, Ankur Deka$^{1}$, Andrew Spielberg$^{2}$, and Matthias M{\"u}ller$^{1}$
\thanks{$^{1}$ S. Brahmbhatt, A. Deka, and M. M{\"u}ller are with Intel Labs. E-mail: {\tt\small \{samarth.manoj.brahmbhatt, ankur.deka, matthias.mueller\}@intel.com}.}%
\thanks{$^{2}$ A. Spielberg is with the Massachusetts Institute of Technology and contributed mostly during an internship at Intel Labs. E-mail: {\tt\small aespielberg@csail.mit.edu}.}%
}

\begin{document}

\makeatletter
\let\@oldmaketitle\@maketitle
\renewcommand{\@maketitle}{\@oldmaketitle
    \begin{figure}[H]
        \setlength{\linewidth}{\textwidth}
        \setlength{\hsize}{\textwidth}
        \centering
        \includegraphics[width=\textwidth]{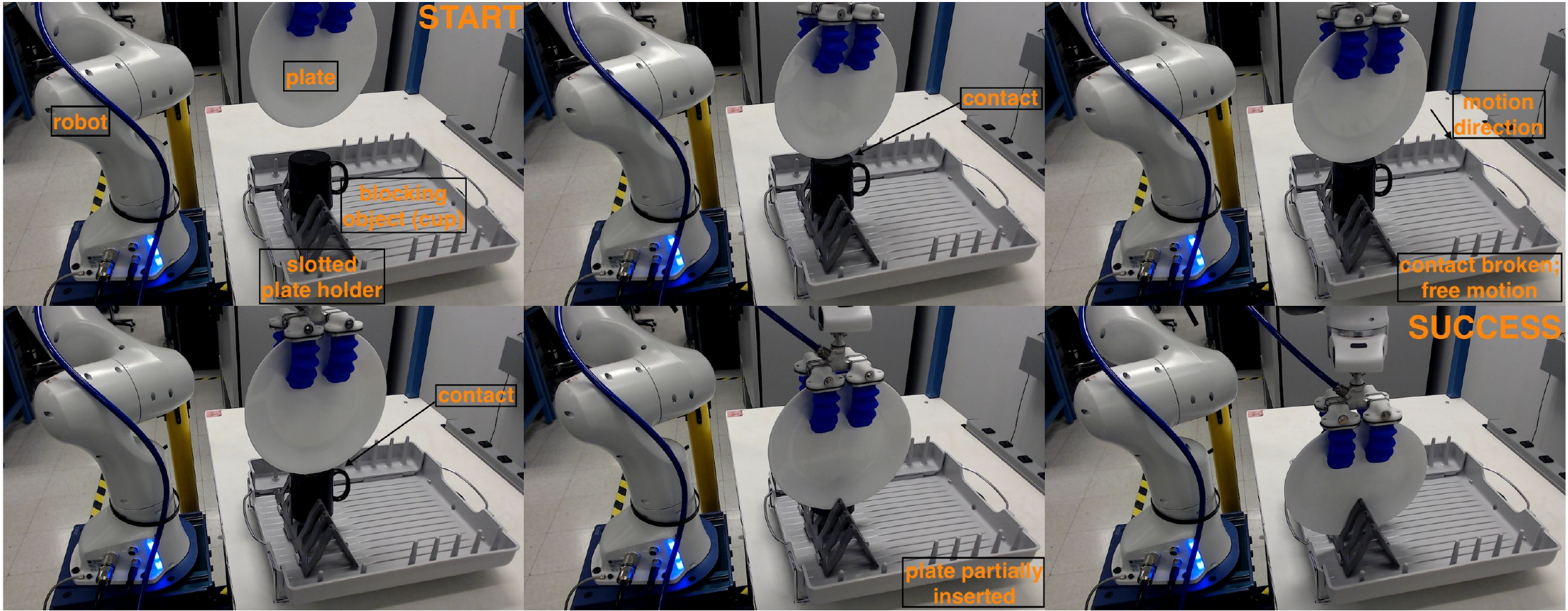}
        \caption{Top-left to bottom-right: sequential frames from a plate insertion episode. The robot is controlled by our plate insertion policy trained entirely in simulation. The policy successfully deals with multiple collisions caused by noise in the target slot location, and an unseen object (a cup) blocking it, and inserts the plate into the neighbouring free slot. Zoom in to see collision details.}
        \label{fig:teaser}
    \end{figure}
}
\makeatother
\maketitle


\begin{abstract}
Humans naturally exploit haptic feedback during contact-rich tasks like loading a dishwasher or stocking a bookshelf. Current robotic systems focus on avoiding unexpected contact, often relying on strategically placed environment sensors. Recently, contact-exploiting manipulation policies have been trained in simulation and deployed on real robots. However, they require some form of real-world adaptation to bridge the sim-to-real gap, which might not be feasible in all scenarios. In this paper we train a contact-exploiting manipulation policy in simulation for the contact-rich household task of loading plates into a slotted holder, which transfers without any fine-tuning to the real robot. We investigate various factors necessary for this zero-shot transfer, like time delay modeling, memory representation, and domain randomization. Our policy transfers with minimal sim-to-real gap and significantly outperforms heuristic and learnt baselines. It also generalizes to plates of different sizes and weights. Demonstration videos and code are available at \url{https://sites.google.com/view/compliant-object-insertion}. 
\end{abstract}



\section{Introduction}\label{sec:intro}
Common household tasks like loading a dishwasher, stocking a bookshelf, or inserting a pot into a coffee maker involve object placement in constrained locations. Humans are able to do these almost effortlessly. For example, when inserting a dish into an empty dishwasher slot, a single glance is enough to approximate the scene geometry and the target pose of the plate. Humans then move the plate towards the slot, handle unexpected collisions along the way, and detect success by leveraging experiential knowledge of how the haptics should feel when the plate is firmly in the slot. Importantly, they do this without access to accurate pose estimation and tracking, and often without live visual feedback. Studies in neurophysics provide evidence that specialized regions in the brain form a ``haptic working memory" from a visual snapshot and use it to plan motions~\cite{maule2015haptic, sathian2016analysis, murata1996parietal}.

Robots need to perform similar object placement tasks, as they are increasingly collaborating with humans in dynamic, unseen, and un-instrumented human environments. However, several challenges currently limit this. The lack of accuracy of or prior access to object and environment 3D models leads to uncertainty in the task state. Unexpected collisions between the object and environment surfaces are therefore very likely to occur, making this a contact-rich task. Contact-exploiting robot motion policies are likely to perform better than contact-avoiding policies in these cases.

\begin{table*}[t]
\centering
\resizebox{\textwidth}{!}{
\begin{tabular}{l c c c c c c c c c c}
  \toprule &
  \ignorecitefornumbering{\citep{broenink1996peg}} &
  \ignorecitefornumbering{\citep{gao2021kpam}} &
  \ignorecitefornumbering{\citep{vecerik2019practical}} &
  \ignorecitefornumbering{\citep{schoettler2020deep}} &
  \ignorecitefornumbering{\citep{beltran2020variable, schoettler2020meta}} &
  \ignorecitefornumbering{\citep{ma2020efficient}} &
  \ignorecitefornumbering{\citep{tang2015learning}} &
  \ignorecitefornumbering{\citep{dong2021tactile}} &
  \ignorecitefornumbering{\citep{lee2020visionandtouch}} &
  ours\\
  \midrule
  Insertion task & peg & peg, USB & peg, clip & plug, USB & plug, gear & pins & peg & boxes & pegs & plates\\
  Learnable policy & \NoX & \YesV & \YesV & \YesV & \YesV & \YesV & \YesV & \YesV & \YesV & \YesV\\
  Generalizes to different geometries & \NoX & \YesV & \NoX & \NoX & \YesV & \NoX & \NoX & \YesV & \YesV & \YesV\\
  \midrule
  Avoids need for unoccluded vision & \YesV & \YesV & \NoX & \NoX & \YesV & \YesV & \YesV & \YesV* & \NoX & \YesV\\
  Avoids need for expert demonstrations & \YesV & \NoX & \NoX & \YesV & \YesV & \NoX & \NoX & \YesV & \YesV & \YesV\\
  Avoids need for real world pre-training & \YesV & \NoX & \YesV & \YesV & \YesV & \YesV & \YesV & \YesV & \NoX & \YesV\\
  Avoids need for real world training / finetuning & \NoX & \YesV & \NoX & \NoX & \NoX & \NoX & \NoX & \NoX & \NoX & \YesV\\
  \bottomrule
\end{tabular}
}  
\caption{Comparison of insertion tasks, policy formulations, and training requirements of the proposed algorithm with related work. *: [22] requires a GelSlim planar tactile sensor.}\label{tab:requirement_comparison}
\end{table*}

However, existing learnt contact-exploiting insertion policies require either real-world expert demonstrations~\ignorecitefornumbering{\citep{tang2015learning, gubbi2020imitation, vecerik2019practical}}, robot training from scratch in the real-world~\ignorecitefornumbering{~\citep{dong2021tactile, schoettler2020deep}}, or real-world adaptation \ie data collection by the robot in the target environment before the policy is fully trained~\ignorecitefornumbering{~\citep{beltran2020variable, schoettler2020meta, lee2020visionandtouch}}. Because this is expensive and potentially unsafe, policies that are trained completely in simulation are desirable. 


In this paper we focus on a common household task: plate insertion into a slotted plate holder (see Figure~\ref{fig:teaser}) when the slot location is noisy, and unseen objects block the target slot. Differently from other works, we investigate how a successful real-world contact-exploiting insertion policy can be trained completely in simulation. The policy is trained using reinforcement learning (RL) considering the delayed-reward nature of the task. The predicted action is passed as a relative 6-DOF motion target to a low-level operational space controller~\ignorecitefornumbering{\cite{osc}}. The observation and reward design encourage contact-awareness under state uncertainty, while the combination of impedance built into the operational space controller and a soft gripper on the real robot result in compliant policy execution. We do not assume access to strategically placed cameras providing live occlusion-free visual feedback. Instead, our policy relies only an approximate target location sensed once \textit{before} the interaction begins. During the interaction, proprioceptive and low-dimensional haptics \ie end-effector wrench features are used to construct policy observations. We also do not assume access to accurate 3D models of the plate or the slotted plate holder.

Real robot experiments demonstrate that our policy successfully inserts the plate into the slot in spite of multiple unexpected collisions, outperforms baselines, and generalizes to plates of different sizes and masses, and even a cup. Through an ablation study, we find that proper memory representation in the policy observation, and incorporating delay between observation capture time and action execution time, significantly increases the transferability of the policy to the real robot. To the best of our knowledge, this is the first non-hardcoded insertion policy trained completely in simulation \ie without pre-training, training, or finetuning with real-world data or demonstrations. To summarize, our contributions are:
\begin{itemize}
\item a haptics-based policy for contact-aware and compliant insertion of plates into a slotted plate holder,
\item demonstration of zero-shot policy transfer to a real robot, comparison to baselines, generalization to insertion of different plates and even a cup, and 
\item an ablation study validating design choices and analyzing their impact on sim-to-real transfer.
\end{itemize}
\section{Related Work}\label{sec:related_work}
Object insertion has been widely studied in robotics in the context of automated industrial part assembly through peg-in-hole insertion~\citep{whitney1987historical, mccallion1979compliant} problems. Approaches range from handcrafted impedance control~\citep{broenink1996peg, tsumugiwa2002variable, park2017compliance}, force control~\citep{stokic1986implementation}, and vision-impedance hybrid control~\citep{morel1998impedance} strategies to learning from human demonstration~\citep{tang2015learning, gubbi2020imitation, stepputtis2022system} and reinforcement learning (RL) with visual, force, and proprioceptive features~\citep{inoue2017deep, beltran2020variable, lee2020visionandtouch}. Some works learn to compose and parameterize handcrafted~\citep{johannsmeier2019framework} or demonstrated~\citep{wu2022primlafd} primitive motions. Recently, RL has also been used for learning key~\citep{pmlr-v155-voigt21a}, clip~\citep{vecerik2019practical}, USB and electrical plug~\citep{schoettler2020deep, schoettler2020meta} insertion policies. We summarize the drawbacks of existing approaches in Table~\ref{tab:requirement_comparison} and elaborate below.

Current RL-based approaches require training or finetuning with the real robot in the target environment e.g.~\citet{lee2020visionandtouch} train a multimodal representation from real-world heuristic policy data,~\citet{vecerik2019practical} and~\citet{ma2020efficient} collect human demonstrations, and~\citet{dong2021tactile},~\citet{schoettler2020deep}, and~\ignorecitefornumbering{\citet{beltran2020variable}} train the RL policy on the real robot. \citet{schoettler2020meta} reduce the amount of real robot training through meta-RL. However, real-world training is expensive to monitor, potentially unsafe, and requires prior access to the target environment. In contrast, we learn the policy only in simulation and transfer it directly to the real robot.

Another common requirement is live visual feedback \eg~\citet{lee2020visionandtouch} and~\citet{schoettler2020deep} use image observations from a camera which is situated such that it can observe the entire interaction without robot body or gripper occlusion. This might not be feasible for a robot operating in an un-instrumented environment.~\citet{gao2021kpam} develop wrist camera image-based oriented keypoint detectors for generating object pose observations, which are used to train an imitation learning peg-hole insertion policy~\cite{morel1998impedance}. In contrast, we rely exclusively on end-effector wrench and proprioceptive features as policy observations. Vision is used to sense the approximate target location only once before the interaction begins, when occlusion is unlikely. Our work is complementary to visual policy methods because our conclusions about increasing the zero-shot transferability of haptics-based policies can be used to augment these methods in cases where strategic camera placement in the environment is acceptable or the grasped object can be tracked under occlusion~\cite{brahmbhatt2015occlusion, grady2021contactopt}.

Our real robot setup, inspired from household robotics, is also different: the target slot is already occupied by an unseen blocking object (see Figure~\ref{fig:environments} and Section~\ref{sec:experiments}), so the policy has to find an empty slot. In contrast, existing approaches \eg~\cite{tang2015learning, gubbi2020imitation, beltran2020variable, lee2020visionandtouch, schoettler2020deep, schoettler2020meta} focus on inserting the object in a single empty slot.

\section{Method} \label{sec:method}

We assume that the object has already been grasped before the policy execution begins. The policy is divided into two stages: an open-loop planner from MoveIt~\citep{moveit0, moveit1} is first used to move the end-effector 12 cm above its approximate target pose (see Section~\ref{sec:experiments} for details). We assume no obstacles are present in this path. The policy then switches to the second stage \ie a learnt controller that handles the contact-rich segment of the task. This is the main contribution of this paper. We describe its details below.

\mypara{Problem formulation}: We model the task as a finite-horizon Markov Decision Process (MDP) with a state space $\mathcal{S}$, continuous action space $\mathcal{A}$, transition dynamics $\mathcal{T}: \mathcal{S} \times \mathcal{A} \rightarrow \mathcal{S}$, discount factor $\gamma \in [0, 1)$, and reward function $R(\mathbf{s}_t, \mathbf{a}_t, \mathbf{a}_{t-1})$. Given a horizon $T$, we aim to learn a policy $\pi: \mathcal{S} \rightarrow \mathcal{P}(\mathcal{A})$ mapping observations to action probabilities $P \in \mathcal{P}$ that maximize the discounted return $J(\pi) = \mathop{\mathbb{E}}_\pi \left[ \sum_{t=0}^{T} \gamma^t R\left( \mathbf{s}_t, \mathbf{a}_t, \mathbf{a}_{t-1} \right) \right]$. $\mathcal{P}$ is the family of unimodal Gaussian PDFs. $\pi$ is parameterized with a 2-layer (256, 256) MLP and is learnt using the Soft Actor-Critic RL algorithm~\citep{sac} containing a 2-layer (256, 256) MLP critic.

\begin{figure}
    \centering
    \includegraphics[width=\columnwidth]{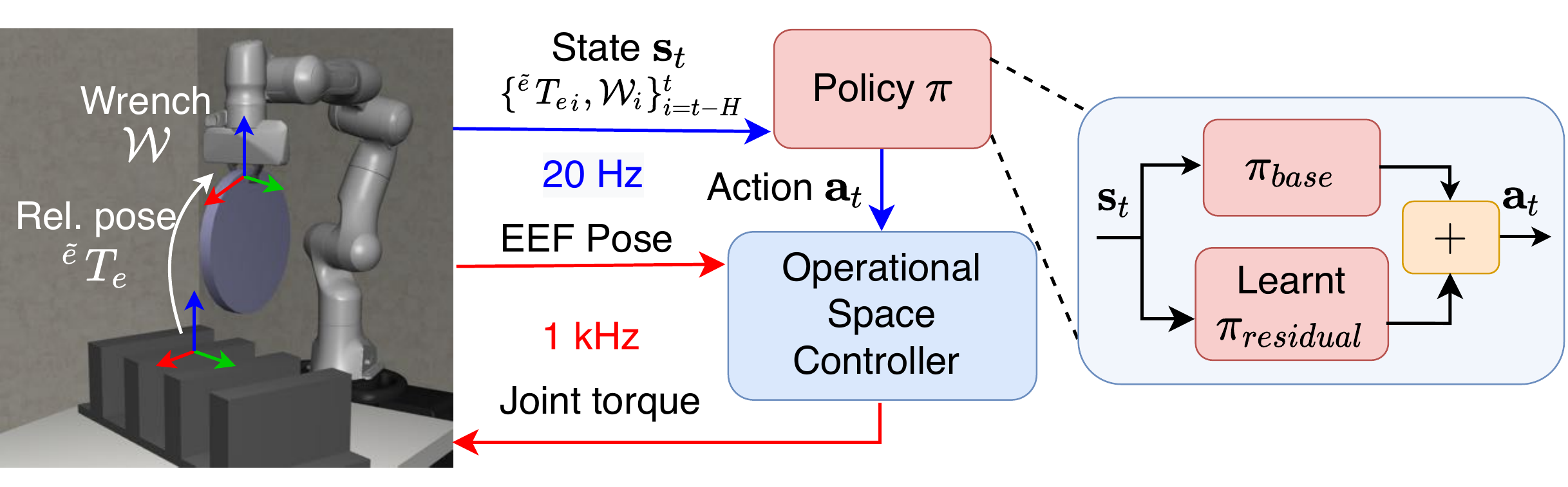}
    \caption{Policy execution architecture.}
    \label{fig:architecture}
\end{figure}

\mypara{Observation}: In the following, $\T{a}{b}$ denotes the 6-DOF pose of coordinate system of body $b$ w.r.t. coordinate system of body $a$. The robot base, target, grasped object, and end-effector are denoted by $b$, $t$, $o$, and $e$ respectively, and the $\bar{~}$ superscript indicates that the body is in its final intended state. The end-effector pose that would place the grasped object at the intended target is thus $\T{b}{\bar{e}} =~\T{b}{t}~\T{t}{\bar{o}}~\T{o}{e}$. However, as mentioned in Section~\ref{sec:intro}, $\T{b}{\bar{e}}$ and the target pose $\T{b}{t}$ are known only approximately in practice. We model this uncertainty by adding 6-DOF noise sampled from a uniform distribution $\mathcal{U}\left[ -\epsilon, \epsilon \right]$ to the perfect $\T{b}{t}$ from simulation, denoting the noisy version by $\Tilde{~}$: $\T{b}{\Tilde{e}} =~\T{b}{\Tilde{t}}~\T{t}{\bar{o}}~\T{o}{e}$. $\T{t}{\bar{o}}$ and $\T{o}{e}$ are assumed constant, even though multiple contacts during the episode keep changing $\T{o}{e}$. This results from the requirement of no object tracking, and means the policy should be robust to in-hand object motion. The observation (see Figure~\ref{fig:architecture}) contains the current end-effector pose w.r.t. the noisy end-effector target pose $\T{\Tilde{e}}{e} =~\T{b}{\Tilde{e}}^{-1}~\T{b}{e}$ (represented with 3 numbers for translation and 3 for axis-angle rotation) and the wrench $\mathcal{W}$ exerted by the end-effector, expressed in the $b$ frame. Gravity, coriolis, and acceleration components are removed from this wrench (see supplementary for details), enabling it to capture object weight and contact forces exclusively. Thus the observation depends on the object mass. \underline{History}: Several relevant quantities like end-effector velocity, contact history, current contact duration etc. require some form of memory. Hence, we stack $\T{\Tilde{e}}{e}$ and $\mathcal{W}$ for $H=8$ steps in the policy observation to increase observability of the true state~\citep{mnih2015human}. We evaluate other memory representations like recurrent policies~\citep{hausknecht2015deep} in Section~\ref{sec:experiments}. See supp. for architecture details.

\mypara{Action}: The policy outputs an action vector $\mathbf{a}_t \in [-1, 1]^6$. The first 3 elements are scaled by 45 cm to get a translation vector, while the last 3 elements are scaled by 45\textdegree~to get an axis-angle rotation vector. The resulting 6-DOF pose is the residual motion target for the end-effector w.r.t. its current pose. Like~\citep{residual_rl}, it is composed with a similarly scaled base motion target that moves the end-effector straight to $\T{b}{\Tilde{e}}$.

\mypara{Low-level controller}: As shown in Figure~\ref{fig:architecture}, the composite motion target is passed to a low-level operational space controller~\citep{osc} running at 1 kHz. A new state vector is captured from the robot after every 50 low-level control steps. The operational space formulation and the end-effector wrench formulation described above decouple the simulated robot dynamics from the real robot dynamics. Because robot manufacturers often have closed-source dynamics APIs and simulators only approximate the dynamics parameters, this is important to reduce the simulation-reality gap. We do not randomize robot dynamics.

\mypara{Reward}: The reward function $R\left(\textbf{s}_t, \textbf{a}_t, \mathbf{a}_{t-1} \right)$ sums the following components: per-step time penalty $R_{time} = -1/T$, object drop penalty (also ends the episode) $R_{drop} = -1.1$, success reward (also ends the episode) $R_{success} = +0.5$, distance to true target penalty $R_{dist} = - \sum_{m \in \{ trans, rot \}} K_{dist}^{(m)} min\left( \lambda_{dist}^{(m)}, \|\T{\Tilde{e}}{e}^{(m)}\| \right)$, and action change penalty for smoothness $R_{\Delta a} = - \sum_{m \in \{ trans, rot \}} K_{\Delta a}^{(m)} min\left( \lambda_{\Delta a}^{(m)}, \|\T{a_{t-1}}{a_t}^{(m)}\| \right)$. $K_{dist}^{(m)}$ and $K_{\Delta a}^{(m)}$ are scale factors, while $\lambda_{dist}^{(m)}$ and $\lambda_{\Delta a}^{(m)}$ are cutoff values. In contrast to some other works like~\citep{lee2020visionandtouch, schoettler2020deep}, we do not need a staged reward function for different phases.

\mypara{Encouraging sim-to-real transfer}: \underline{$\T{b}{t}$ noise}: In initial experiments without noise added to $\T{b}{t}$ we found that the policy overfits by remembering the obstacle geometry and learning a $\T{b}{\bar{e}}$ trajectory that avoids all collisions. However, such a policy fails on the real robot because the observations are noisy and collisions become inevitable. Adding 6-DOF noise $\mathcal{U}\left[ -\epsilon, \epsilon \right]$ to $\T{b}{t}$ as described above forces collisions, because the true target pose is no longer observable by the policy. This allows us to achieve the objective of learning collision-friendly policies. We increase the $\T{b}{t}$ noise parameter $\epsilon$ in a piecewise linear curriculum. \underline{Partial insertion initialization}: But this also significantly reduces the probability of the object reaching its intended target pose with purely random exploration. Hence, following reverse curriculum generation~\cite{florensa2017reverse}, we initialize the robot in 50\% of the episodes in poses such that the object is already partially inserted. \underline{Policy inference delay}: In a simulator, time stops during policy inference and round-trip communication of state and action from the environment to the policy. This is not true for the real world~\cite{sandha2021sim2real, andrychowicz2020learning}. Hence we randomize this delay in the range of $[7, 13]$ low-level control steps (based on delay in the real robot setup) and found this to be significant for sim-to-real transfer (see~Table \ref{tab:success_rates}).

\mypara{Policy learning}: Contact simulation and the randomizations described above increase the simulation and environment reset complexity. Hence we learn the policy with the off-policy Soft Actor Critic (SAC)~\citep{sac} algorithm, because it is more sample efficient than on-policy algorithms~\cite{schulman2017proximal}. We seed the replay buffer with 500 episodes of transitions collected with a random policy to encourage exploration. `Success' is detected when the object pose is below a height threshold and the tabletop in-plane coordinates (X and Y) are within $d_{success}$ distance of any slot center. To dampen in-place oscillations once the plate is inserted in the slot, we increased the number of policy steps in the `success' state required to end the episode from 1 to 10, and $d_{success}$ to cover the entire slot width.

\mypara{Implementation details}: We use the simulation environment and operational space controller implementations in Robosuite~\citep{robosuite2020}, which uses the MuJoCo physics engine~\citep{mujoco}. Training is performed using tf-agents~\citep{TFAgents} and Reverb~\citep{reverb}. While the Franka Emika Panda robot~\citep{panda_robot} is used for both simulation and real-world experiments, the former use the Panda rigid gripper and the latter use a Soft Robotics Inc.\ mGrip gripper~\citep{soft_gripper} (see Fig.~\ref{fig:environments}). We found that in practice the mGrip's high friction and balance of firmness and compliance allows it to hold thin objects like plates much better than the Panda rigid gripper. As our experiments validate, the policy trained in simulation transfers well to the real world in spite of this difference. Modeling the mGrip gripper in simulation for potentially more accurate wrench observations is out of the scope of this paper. See the supplementary material for values of all hyperparameters mentioned in this section. We trained the learnable policies on a cluster with Intel(R) Xeon(R) Platinum 8180 CPU @ 2.50 GHz, 112 cores and 1 TB RAM. Training one policy took 14 hours.
\section{Experiments}\label{sec:experiments}
\begin{figure}[ht]
    \centering
    \includegraphics[width=\linewidth]{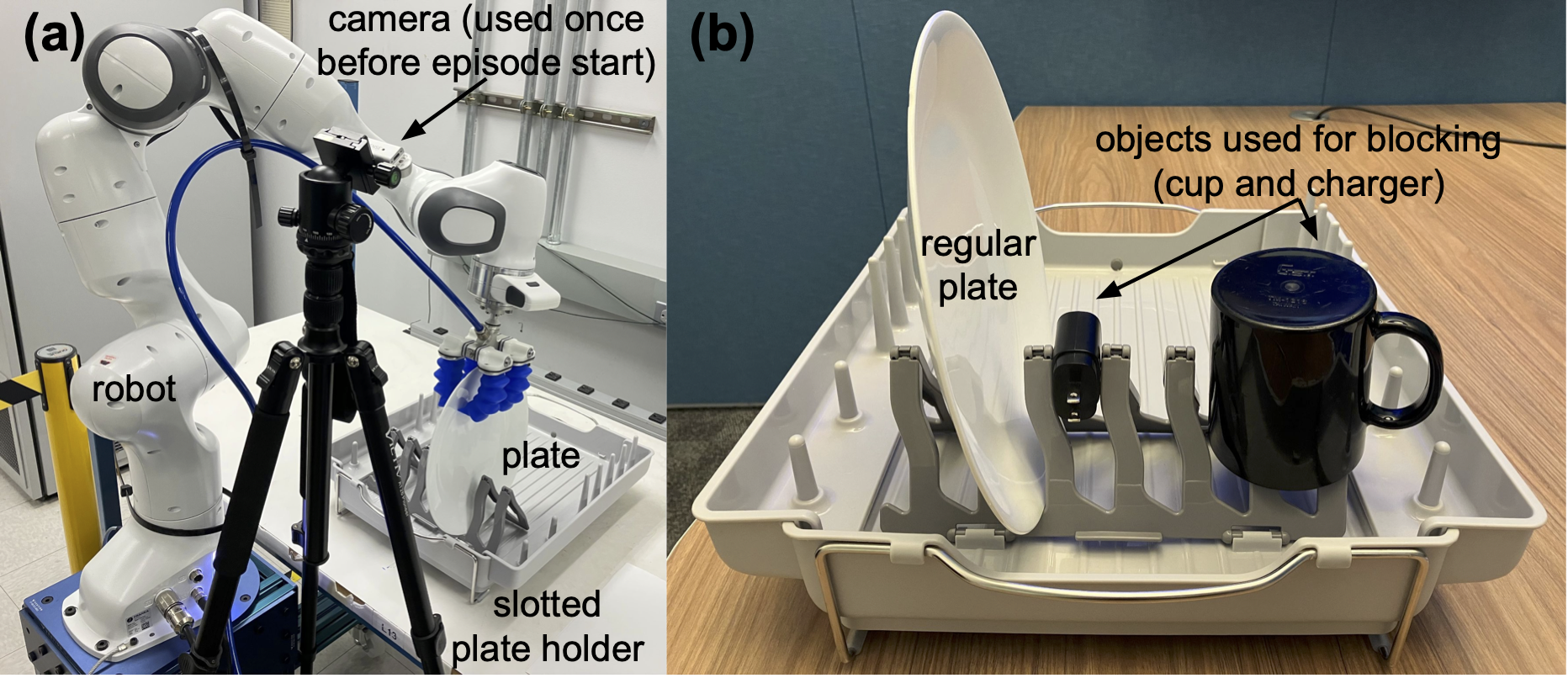}
    \caption{\textbf{(a)}: Real robot policy execution setup. \textbf{(b)}: Detailed view of the main plate, plate holder slots, and the unseen blocking objects - cup and charger.}
    \label{fig:environments}
\end{figure}

\mypara{Real robot experiments evaluation protocol}: The episode starts with the end-effector at a random pose in a cuboidal region of dimensions 20 $\times$ 40 $\times$ 15 cm above the central slot of the plate holder. We assume that the object has been grasped before the episode starts. In simulation, this is done by applying gravity compensation to the object held at the appropriate graspable pose, while in the real world an operator holds it between the gripper fingers while the grip closes. The operator holds an ArUco~\cite{aruco0, aruco1} marker above the target slot, estimating the end effector location for the plate to be fully inserted. The marker pose is detected by the shoulder camera and treated as the noisy target end effector pose \wrt robot base $\T{b}{\Tilde{e}}$ used to calculate observations (see Section~\ref{sec:method}). An episode succeeds when the plate is inserted in any slot (determined by thresholding the end-effector height and medial axis error \wrt $\T{b}{\Tilde{e}}$) and is kept there for 3 seconds. We release the grasp and raise the arm upon detecting success. It fails when the plate is kept stationary at a location other than the slots (including jamming with high force), or the robot hits torque or joint angle limits. The policy trained in simulation is deployed directly on the real robot without adaptation. As seen in Figure~\ref{fig:environments}, the plate holder has 6 slots. We perform 4 trials for each slot and compute the success rate separately across slots for the first, second, third, and fourth trial to get 4 success rate values. We report the mean and standard deviation of these for each method. 

\mypara{Blocking objects}: As explained in Section~\ref{sec:method}, noise is added in the observations during simulation training to force the policy to learn collision-friendly behaviours. However, the real plate holder slots are much more ergonomically designed than the cuboidal geometries used in simulation (see Figure~\ref{fig:environments}). Even though the clearance between the plate and slot widths is 2.5 mm in the real setup, a policy that centers above the slot and moves downward without drift is likely to succeed trivially. Hence, to rigorously test the learnt policies' collision behaviour we place a blocking object e.g. a cup or a phone charger in the target slot. This tests the policy's ability to handle unexpected collisions, and better reflects real-world application conditions like a mobile manipulator attempting to load a partially loaded dishwasher.

\begin{figure}
\centering
\includegraphics[width=\linewidth]{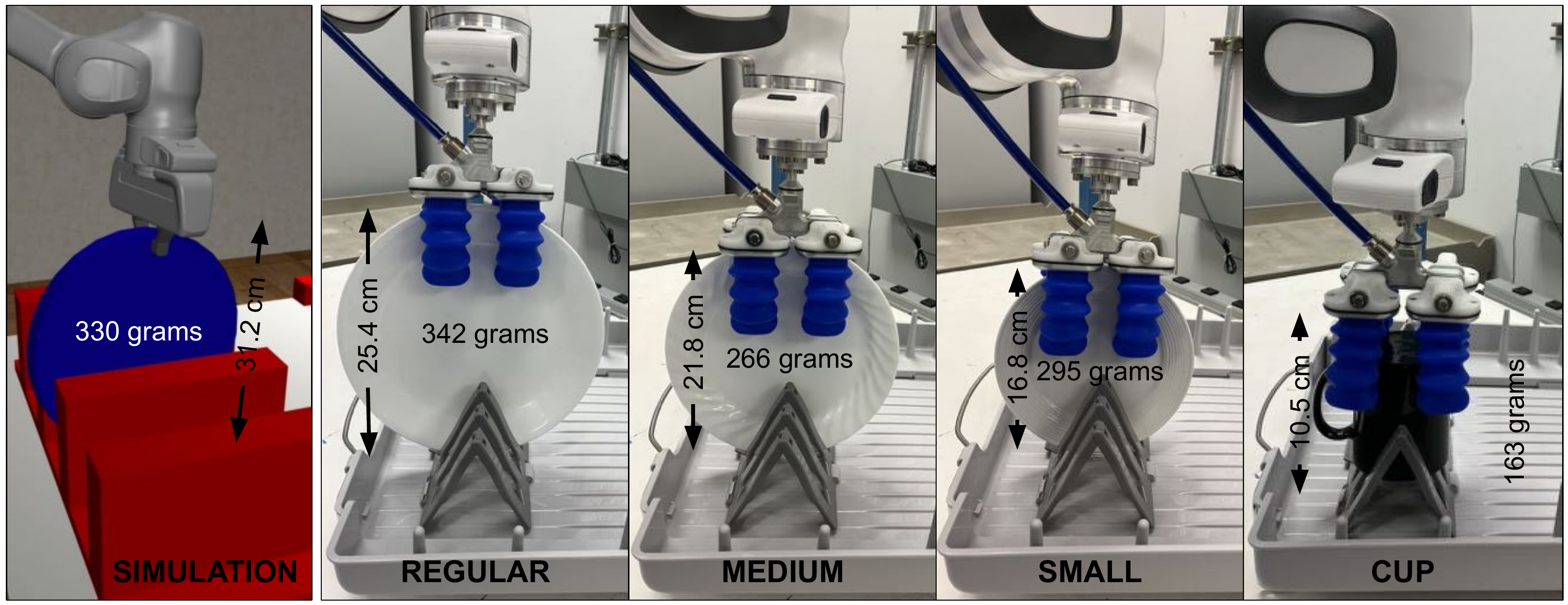}
\caption{Characteristics of the simulated plate used for training, and plates and cup used for real robot experiments, showing the highly approximate nature of simulation geometry. The ablation study (Figure~\ref{fig:ablation}) is done with the ``regular" plate. Note also the gripper differences.}
\label{fig:plates}
\end{figure}

\mypara{Differences from simulation}: Figure~\ref{fig:plates} shows how the simulated plate and slotted holder geometries are very rough approximations of their real counterparts. Other approximations include the ideal plate pose in its target state w.r.t. the slotted holder $\T{t}{\bar{o}}$, the pose of the end-effector w.r.t. the grasped plate $\T{o}{e}$, soft gripper simulated as a rigid gripper with spring-loaded joints, and blocking the target slot with a cup or charger. Our proposed algorithm yields high-performance policies despite these large differences.

\mypara{Baselines}: We compare our policy to the following baselines also implemented at the low-level through the same operational space controller as ours:
\begin{itemize}
    \item \texttt{straight-down}: The end-effector moves downwards indefinitely, implemented by setting the operational space controller's target to $[0, 0, -0.05]$ cm.
    \item \texttt{random-search}~\citep{marvel2018multi}: The end-effector moves to a point randomly sampled from a horizontal square of side length 5 cm, and then downwards till a contact force of 3 N is experienced. If the episode does not succeed in this pose, it moves upwards, samples another point from the square, and repeats.
    \item \citet{lee2020visionandtouch} \texttt{no vision}: While the full vision-based policy from~\citep{lee2020visionandtouch} is not comparable to this work, its ``no vision" variant provides a useful baseline for network architecture. It has separate branches for proprioception and wrench observations, and uses temporal convolution to represent history. We train the policy neural network with all the randomizations described in Section~\ref{sec:method} directly on our RL objective instead of the auxiliary pre-training objectives from \citet{lee2020visionandtouch} to avoid training with the real robot. The original architecture has a large number of learnable parameters and does not perform well in simulation (17.0 $\pm$ 13.1\% success rate). Hence we compare to a version with smaller 16-d wrench and proprioception feature vectors.
\end{itemize}

\begin{figure*}[ht]
\centering
\includegraphics[width=\textwidth]{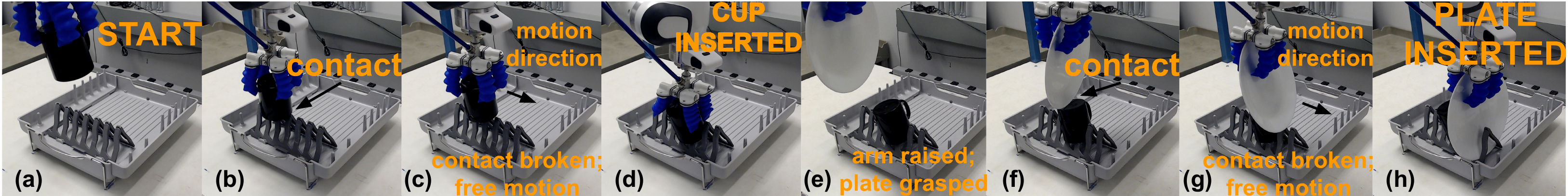}
\caption{Left to right: Sequential video frames showing multiple object insertion and the generalization ability of our policy. Our policy, which is trained in simulation to insert a plate (Figure~\ref{fig:plates}) into a slot whose location is only approximately known, is deployed on the real robot without any real-world adaptation. \textbf{(a)}-\textbf{(d)}: It first inserts a cup, a completely unseen object. \textbf{(e)}: When success is detected, the grasp is released, arm raised, and the ``regular" plate is grasped. The same policy is commanded to insert the plate in the slot blocked by the cup. \textbf{(f)-(h)}: It inserts the plate in the next free slot. Both episodes exhibit the collision-exploiting behaviour learnt by our policy. Zoom in to see collision details. Full video in the supplementary material.}
\label{fig:cup_plate}
\end{figure*}

\mypara{Comparison to baselines}: Table~\ref{tab:success_rates} compares our full policy with the three baseline policies in terms of real-world success rate. Our full policy significantly outperforms them and exhibits the ability to deal with multiple collisions and finish the task (see Figures~\ref{fig:teaser},~\ref{fig:cup_plate}). Qualitatively, the \texttt{straight-down} baseline is sensitive to noise in the target pose, while the \texttt{random-search} baseline is sensitive to yaw error in the target pose, which causes the plate to contact the slot sides when it is only partially inserted, triggering an upward motion. The \texttt{no vision} policy~\citep{lee2020visionandtouch} performs better in simulation with smaller feature vectors but overfits, because the real robot performance is significantly lower. Qualitatively, it often keeps retrying at the same location after recovering from collisions.

\begin{figure}
\centering
    \includegraphics[width=\linewidth]{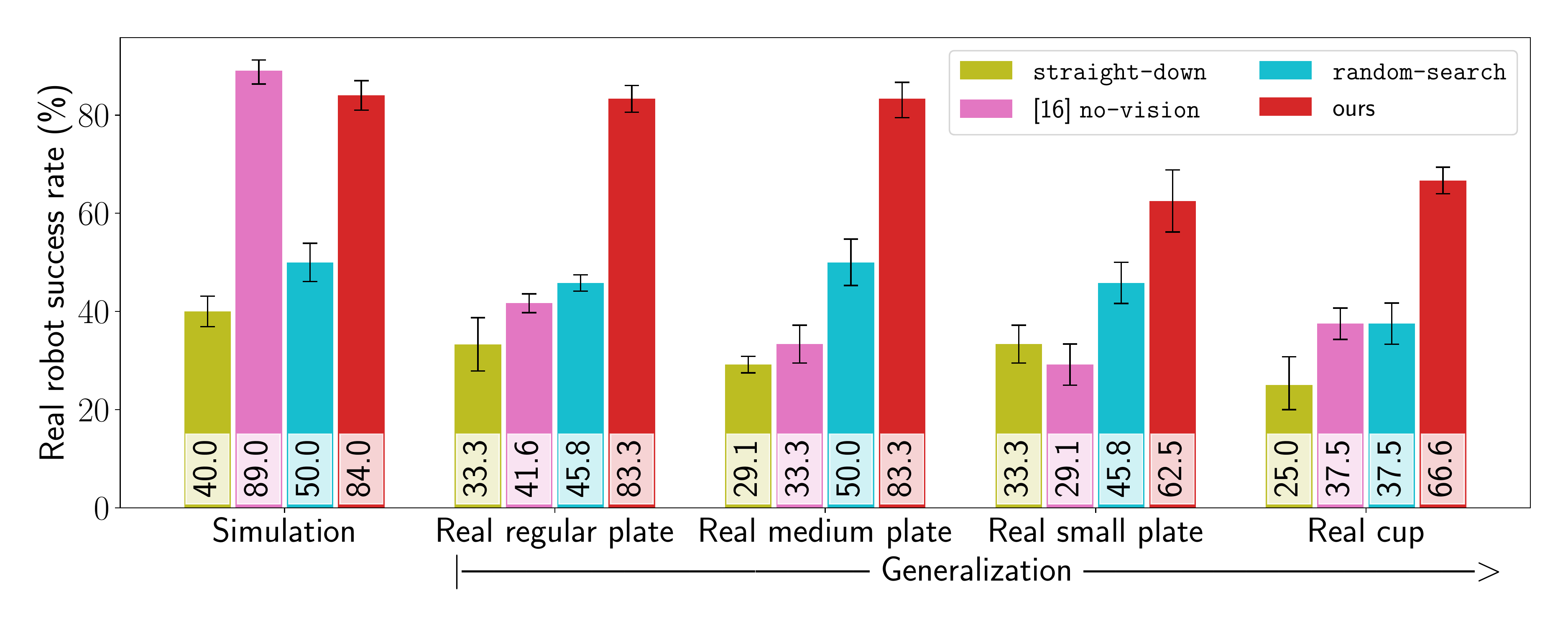}
\caption{\textbf{Generalization}: Real robot performance for different plates shown in Figure~\ref{fig:plates} (error bars show 20\% std. dev.). Simulation performance is also shown for reference. Our algorithm consistently outperforms baselines.}
\label{fig:generalization}
\end{figure}

\mypara{Generalization}: All policies are trained in simulation with a perfectly cylindrical plate, while the real robot experiments are done with three plates of varying sizes, shapes, and weights, all different from the simulated plate (see Figure~\ref{fig:plates}). As shown in Figure~\ref{fig:generalization}, our policy consistently outperforms baselines across plates. It also generalizes with almost the same performance from simulation to the ``regular" and ``medium" plates. We also used the same policy to insert the ``small" plate and the cup shown in Figure~\ref{fig:plates}. Even though their size and geometry differs significantly from the training simulated plate, we found that our policy performs reasonably well after adding a constant offset to the Z translation component of $\T{\Tilde{e}}{e}$ observations to compensate for the height difference. Figure~\ref{fig:cup_plate} shows a multiple object insertion application with our policy, where the cup is inserted first and the plate, targeted to the slot covered by the cup, is then inserted in the next free slot.

\begin{figure}
\centering
    \includegraphics[width=\linewidth]{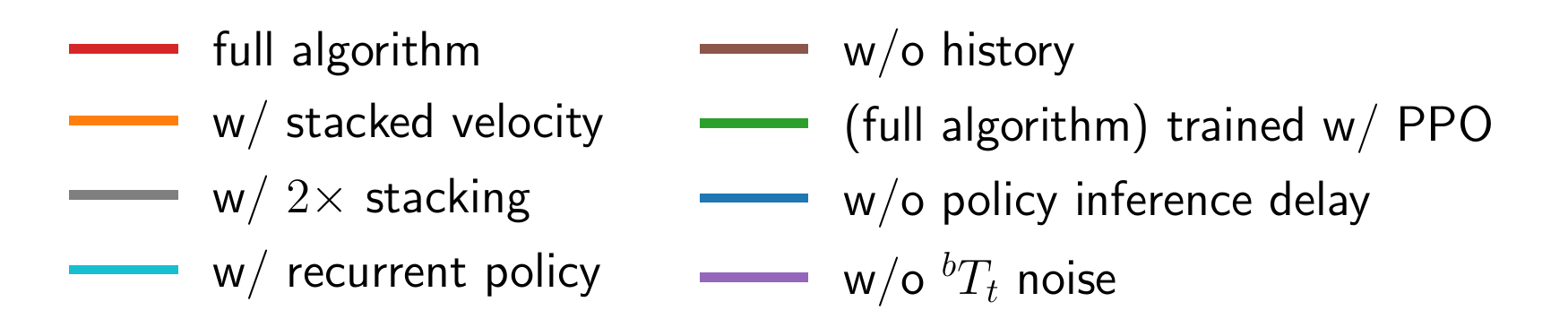}
    \includegraphics[width=\linewidth]{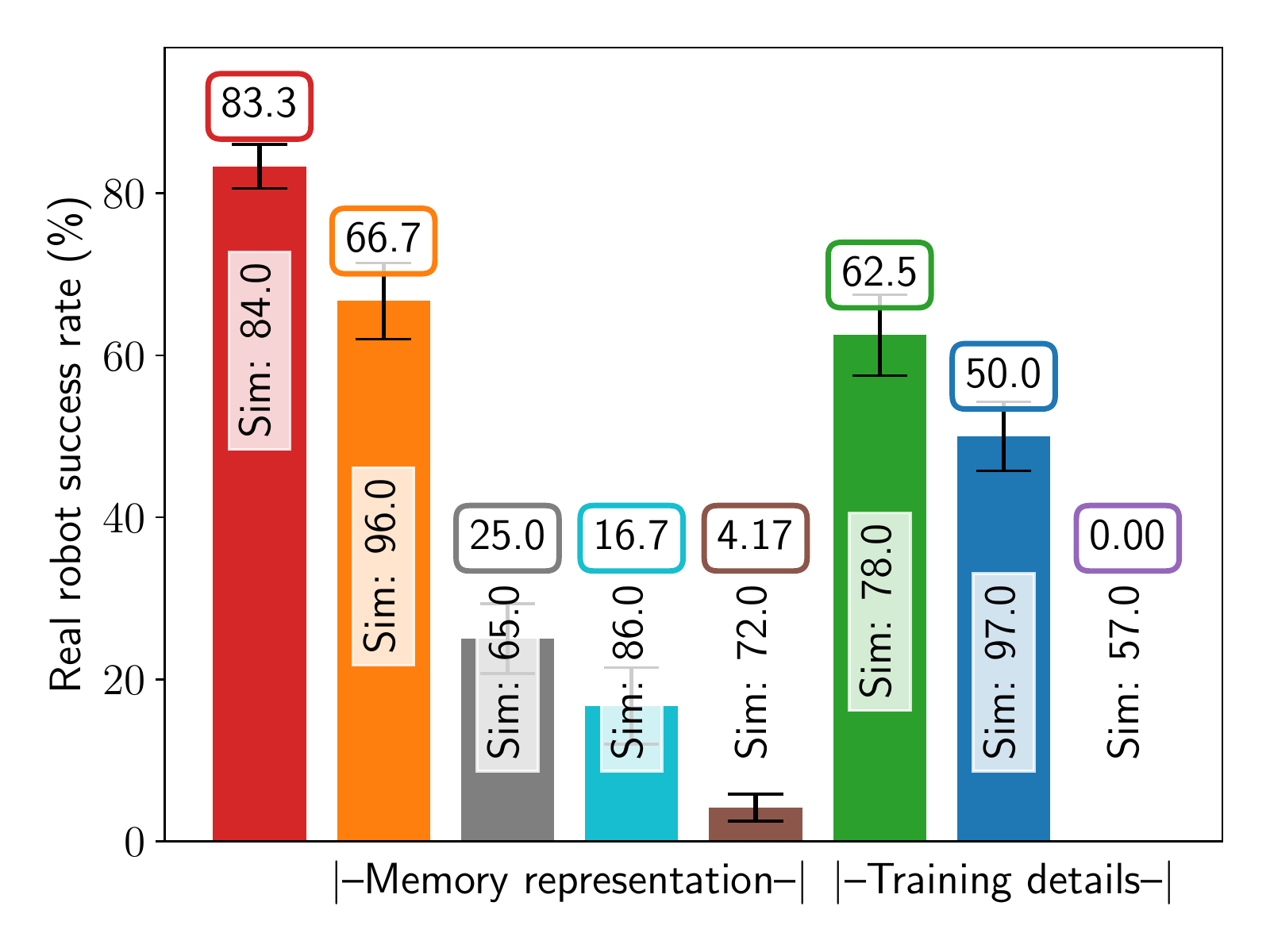}
\caption{\textbf{Ablation study}: Impact of various design choices. Bar height indicates real robot success rate, while simulation success rates are mentioned inside the bars (error bars show 20\% std. dev.). Our full algorithm, which represents memory with $H=8$ stacked observations, performs best. The importance of training with SAC, policy inference delay, and $\T{b}{t}$ noise in observations is also shown.}
\label{fig:ablation}
\end{figure}

\mypara{Ablation study}: Figure~\ref{fig:ablation} shows the simulation and real robot performance of various ablated versions of our full policies.  All policies are evaluated under the same `full policy' conditions with the ``regular" plate i.e. full-scale $\T{b}{t}$ noise in observations and non-zero policy inference delay. \underline{History representation}: The experiments with the real robot show that observation stacking, incorporating recurrent units in the policy network, and including gripper velocity w.r.t. goal in observations all improve true state observability compared to no history, because they result in higher success rate. Observation stacking performs best. However, the policies with $H=16$ and explicit gripper velocity significantly underperform our main policy ($H=8$), probably because a larger observation vector linearly increases the number of learnable parameters. \underline{Sim-to-real transfer}: Training with a history of stacked observations, $\T{b}{t}$ noise in observations, and a non-zero policy inference delay are crucial design choices for ensuring the policy's transferability to the real robot. \underline{Training}: In addition (not shown in Fig.~\ref{fig:ablation}), the policy completely failed in simulation training without partial insertion initialization and the residual action formulation.

\begin{table}
\centering
\resizebox{1\linewidth}{!}{
\begin{tabular}{l c c}
  \toprule
  \multirow{2}{*}{\textbf{Method}} & \textbf{Sim. success rate (\%)} & \textbf{Real success rate (\%)}\\
  & \multicolumn{2}{c}{mean over 4 trials $\pm$ std. dev.}\\
  \midrule
  \texttt{straight-down} & 40.0 $\pm$ 15.5 & 33.3 $\pm$ 27.2\\
  \citep{lee2020visionandtouch} \texttt{no-vision} & \textbf{89.0} $\pm$ 13.4 & 41.7 $\pm$ 9.62\\
  \texttt{random-search} & 50.0 $\pm$ 19.5 & 45.8 $\pm$ 8.33\\
  ours & 84.0 $\pm$ 15.0 & \textbf{83.3} $\pm$ 13.6\\
  \bottomrule
\end{tabular}
}  
\caption{Plate insertion performance in simulation and real-world. Our full algorithm significantly outperforms other methods in real-world performance, and has the smallest sim-to-real gap.}\label{tab:success_rates}
\end{table}

\begin{figure}
\centering
\includegraphics[width=\linewidth]{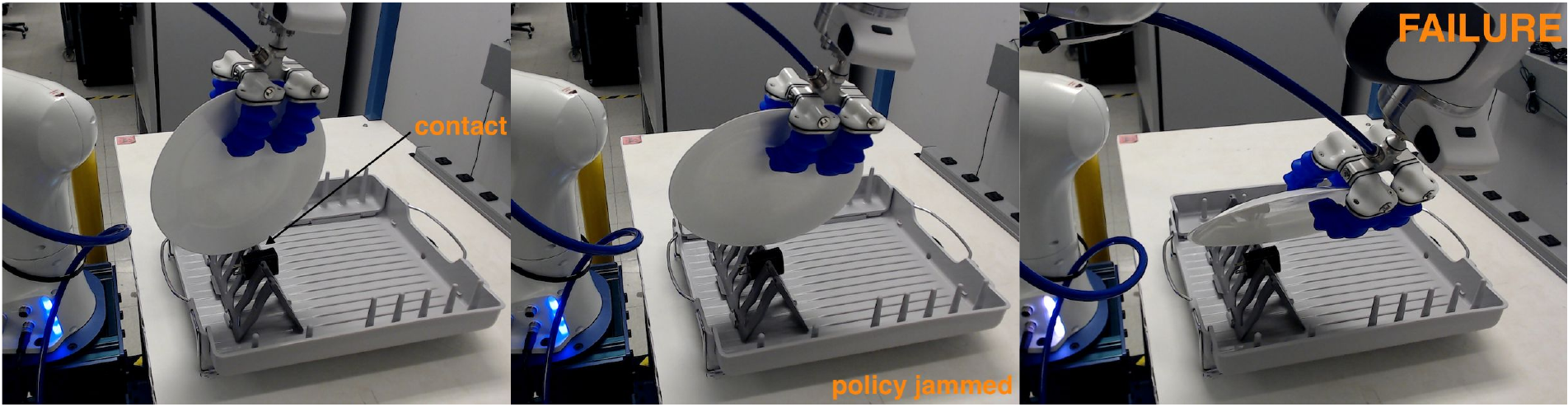}
\caption{Left to right: Sequential video frames showing a failure case of our policy where it gets jammed in a collision and cannot recover. Zoom in to see collision details. Full video in the supplementary material.}
\label{fig:failure}
\end{figure}


\mypara{Qualitative behaviour}: The policy trained without the policy inference delay has very limited responsiveness to collisions, while that trained without observation history drifts and tends to converge to a set of fixed joint positions. Finally, the policy trained without noise cannot differentiate between collisions with slot tops and bottoms and hence lifts the plate out even after successfully inserting it sometimes. While our full policy achieves almost identical performance in simulation and on the real robot, the sim-to-real gap increases significantly for the ablated models, highlighting the importance of each component. Multiple policy executions are shown in the supplementary video.

\mypara{Limitations}: Currently the policy does not perform well when there is a large horizontal gap between the end-effector pose at the start of stage 2 and the target pose $\T{b}{\Tilde{e}}$, and when the base motor is rotated far from its center, even though it is trained in simulation under these conditions. We hypothesize that this is caused by imperfectly identified inertial parameters used in the operational space controller.

\section{Conclusion}\label{sec:conclusion}
We presented a robot learning system for contact-rich object insertion that uses end-effector wrench and proprioception feedback, and a noisy insertion target pose. In contrast to other works, the policy is trained exclusively in simulation and transfers directly to the real robot. We demonstrate the capability of our approach on the task of inserting a plate into a slotted plate holder and generalizing to plates of different sizes and masses, through an ablation study and comparisons to other baseline methods.
We further show that our policy can also insert a cup, an object with a completely different geometry than the training simulation plate. The ability of this system to deal with uncertainty in the insertion target pose can potentially be useful in mobile manipulation use-cases, where robot base motion is a significant source of manipulation uncertainty. We hope to integrate this system with a mobile manipulator in the future.

\mypara{Acknowledgements}: We would like to thank Benoit Landry for help with real robot controller implementation. We would also like to thank Vladlen Koltun for contributions to the problem statement definition and feedback on real robot experiments.


\section*{Supplementary Material}

\subsection{Wrench calculation for policy observations}
MuJoCo has 3 types of soft constraints: equality, friction loss, and contact. However, it reports the total joint torque arising from all constraints $\tau_{constraint}$, not separately. For the policy observations we need to calculate the joint torque contribution of just the contact constraints. We calculate it as $\tau_{contact} = \tau_{constraint} - \sum_c J_c^T \left( \theta \right) F_c$, where $J_c \left( \theta \right)$ is the constraint Jacobian for constraint $c$ (dependent on joint positions $\theta$), and $F_c$ is the task-space force of constraint $c$ (which MuJoCo does report separately for all constraints). Here $c$ excludes contact constraints. Next, wrench exerted by the robot at end-effector frame $K$ is calculated as $\mathcal{W} = -J_K^{-T} \left( \theta \right) \tau_{contact}$, where $J_K^{-T}$ is the pseudo-inverse of $J_K^{T}$ and the negative sign converts torque exerted \textit{on} the robot by contact constraints to torque exerted \textit{by} the robot.

\subsection{Recurrent policy architecture}
The recurrent policy (See Table 2 in the main paper) uses the architecture \texttt{MLP(256) - LSTM(256) - MLP(256} for the actor policy $\pi$ and \texttt{MLP(256) - LSTM(256) - MLP(200) - MLP(200)} for the critic.
\subsection{Hyperparameters}\label{sec:hyperparams}
Table~\ref{tab:hyperparams} mentions the values of hyperparameters used for training policies.

\begin{table*}
\centering
\begin{tabular}{l c}
  \toprule
  \textbf{Hyperparameter} & \textbf{Value}\\
  \midrule
  \multicolumn{2}{c}{\textbf{Environment}}\\
  Horizon $T$ & 128\\
  Gap between slots in simulation & 10 cm\\
  Number of slots in simulation & 3\\
  $\T{b}{t}$ noise $\epsilon_T$ & 5 cm, 5\textdegree\\
  Number of steps concatenated for observation history $H$ & 8\\
  Max. distance from target for success & 2 cm, 10\textdegree\\
  \midrule
  \multicolumn{2}{c}{\textbf{Reward}}\\
  Time penalty $R_{time}$ & $-1/128$\\
  Plate drop penalty $R_{drop}$ & $-1.1$\\
  Success reward & $+0.5$\\
  $K_{dist}^{\left( trans \right)}$, $K_{dist}^{\left( rot \right)}$ & $8.59 \times 10^{-3}$, $8.21 \times 10^{-3}$\\
  $K_{\Delta a}^{\left( trans \right)}$, $K_{\Delta a}^{\left( rot \right)}$ & $8.59 \times 10^{-3}$, $8.21 \times 10^{-3}$\\
  $\lambda_{dist}^{\left( trans \right)}$, $\lambda_{dist}^{\left( rot \right)}$ & 50 cm, 30\textdegree\\
  $\lambda_{\Delta a}^{\left( trans \right)}$, $\lambda_{\Delta a}^{\left( rot \right)}$ & 50 cm, 30\textdegree\\
  \midrule
  \multicolumn{2}{c}{\textbf{Low-level operational space controller}}\\
  Proportional gain & 150.0\\
  Damping ratio & 1.0\\
  Translation error scaling factor & 0.05\\
  Rotation error scaling factor & 0.5\\
  Frequency & 1 KHz\\
  \midrule
  \multicolumn{2}{c}{\textbf{High-level policy controller}}\\
  Scaling factor for translation target predicted in $[-1, 1]$ & 45 cm\\
  Scaling factor for rotation target predicted in $[-1, 1]$ & 50\textdegree\\
  Frequency & 20 Hz\\
  Low-level controller steps per policy step randomization range & $[7, 13]$\\
  \midrule
  \multicolumn{2}{c}{\textbf{Soft actor-critic algorithm}}\\
  Batch size & 256\\
  Discount factor $\gamma$ & 0.99\\
  Initial $\alpha$ & $e$\\
  Actor, critic, and $\alpha$ learning rate & $3 \times 10^{-4}$\\
  Training iterations & 64 K\\
  Policy updates per training iteration & 50\\
  Experience collected per training iteration & 140 steps approx.\\
  Parallel experience collection workers & 20\\
  Replay buffer size & 1024 K steps\\
  Experience collected with random policy before training starts & 64 K steps\\
  \bottomrule
\end{tabular}
\caption{Hyperparameters}\label{tab:hyperparams}
\end{table*}
\FloatBarrier
\bibliographystyle{IEEEtran}
\bibliography{IEEEabrv,references}

\begin{thebibliography}{10}
\providecommand{\url}[1]{#1}
\csname url@rmstyle\endcsname
\providecommand{\newblock}{\relax}
\providecommand{\bibinfo}[2]{#2}
\providecommand\BIBentrySTDinterwordspacing{\spaceskip=0pt\relax}
\providecommand\BIBentryALTinterwordstretchfactor{4}
\providecommand\BIBentryALTinterwordspacing{\spaceskip=\fontdimen2\font plus
\BIBentryALTinterwordstretchfactor\fontdimen3\font minus
  \fontdimen4\font\relax}
\providecommand\BIBforeignlanguage[2]{{%
\expandafter\ifx\csname l@#1\endcsname\relax
\typeout{** WARNING: IEEEtran.bst: No hyphenation pattern has been}%
\typeout{** loaded for the language `#1'. Using the pattern for}%
\typeout{** the default language instead.}%
\else
\language=\csname l@#1\endcsname
\fi
#2}}

\bibitem{maule2015haptic}
F.~Maule, G.~Barchiesi, T.~Brochier, and L.~Cattaneo, ``Haptic working memory
  for grasping: the role of the parietal operculum,'' \emph{Cerebral Cortex},
  vol.~25, no.~2, pp. 528--537, 2015.

\bibitem{sathian2016analysis}
K.~Sathian, ``Analysis of haptic information in the cerebral cortex,''
  \emph{Journal of Neurophysiology}, vol. 116, no.~4, pp. 1795--1806, 2016.

\bibitem{murata1996parietal}
A.~Murata, V.~Gallese, M.~Kaseda, and H.~Sakata, ``Parietal neurons related to
  memory-guided hand manipulation,'' \emph{Journal of Neurophysiology},
  vol.~75, no.~5, pp. 2180--2186, 1996.

\bibitem{whitney1987historical}
D.~E. Whitney, ``Historical perspective and state of the art in robot force
  control,'' \emph{The International Journal of Robotics Research}, vol.~6,
  no.~1, pp. 3--14, 1987.

\bibitem{mccallion1979compliant}
H.~McCallion, G.~Johnson, and D.~Pham, ``A compliant device for inserting a peg
  in a hole,'' \emph{Industrial Robot: An International Journal}, 1979.

\bibitem{broenink1996peg}
J.~F. Broenink and M.~L. Tiernego, ``Peg-in-hole assembly using impedance
  control with a 6 dof robot,'' in \emph{Proceedings of the 8th European
  Simulation Symposium}.\hskip 1em plus 0.5em minus 0.4em\relax Citeseer, 1996,
  pp. 504--508.

\bibitem{tsumugiwa2002variable}
T.~Tsumugiwa, R.~Yokogawa, and K.~Hara, ``Variable impedance control with
  virtual stiffness for human-robot cooperative peg-in-hole task,'' in
  \emph{IEEE/RSJ international conference on intelligent robots and systems},
  vol.~2.\hskip 1em plus 0.5em minus 0.4em\relax IEEE, 2002, pp. 1075--1081.

\bibitem{park2017compliance}
H.~Park, J.~Park, D.-H. Lee, J.-H. Park, M.-H. Baeg, and J.-H. Bae,
  ``Compliance-based robotic peg-in-hole assembly strategy without force
  feedback,'' \emph{IEEE Transactions on Industrial Electronics}, vol.~64,
  no.~8, pp. 6299--6309, 2017.

\bibitem{stokic1986implementation}
D.~Stokic, M.~vnkobratovic, and D.~Hristic, ``Implementation of force feedback
  in manipulation robots,'' \emph{The International journal of robotics
  research}, vol.~5, no.~1, pp. 66--76, 1986.

\bibitem{morel1998impedance}
G.~Morel, E.~Malis, and S.~Boudet, ``Impedance based combination of visual and
  force control,'' in \emph{Proceedings. 1998 IEEE International Conference on
  Robotics and Automation (Cat. No.98CH36146)}, vol.~2, 1998, pp. 1743--1748
  vol.2.

\bibitem{tang2015learning}
T.~Tang, H.-C. Lin, and M.~Tomizuka, ``A learning-based framework for robot
  peg-hole-insertion,'' in \emph{Dynamic Systems and Control Conference}, vol.
  57250.\hskip 1em plus 0.5em minus 0.4em\relax American Society of Mechanical
  Engineers, 2015, p. V002T27A002.

\bibitem{gubbi2020imitation}
S.~Gubbi, S.~Kolathaya, and B.~Amrutur, ``Imitation learning for high precision
  peg-in-hole tasks,'' in \emph{2020 6th International Conference on Control,
  Automation and Robotics (ICCAR)}, 2020, pp. 368--372.

\bibitem{stepputtis2022system}
S.~Stepputtis, M.~Bandari, S.~Schaal, and H.~B. Amor, ``A system for imitation
  learning of contact-rich bimanual manipulation policies,'' in \emph{2022
  IEEE/RSJ International Conference on Intelligent Robots and Systems
  (IROS)}.\hskip 1em plus 0.5em minus 0.4em\relax IEEE, 2022, pp.
  11\,810--11\,817.

\bibitem{inoue2017deep}
T.~Inoue, G.~De~Magistris, A.~Munawar, T.~Yokoya, and R.~Tachibana, ``Deep
  reinforcement learning for high precision assembly tasks,'' in \emph{2017
  IEEE/RSJ International Conference on Intelligent Robots and Systems (IROS)},
  2017, pp. 819--825.

\bibitem{beltran2020variable}
C.~C. Beltran-Hernandez, D.~Petit, I.~G. Ramirez-Alpizar, and K.~Harada,
  ``Variable compliance control for robotic peg-in-hole assembly: A
  deep-reinforcement-learning approach,'' \emph{Applied Sciences}, vol.~10,
  no.~19, p. 6923, 2020.

\bibitem{lee2020visionandtouch}
M.~A. Lee, Y.~Zhu, P.~Zachares, M.~Tan, K.~Srinivasan, S.~Savarese, L.~Fei-Fei,
  A.~Garg, and J.~Bohg, ``Making sense of vision and touch: Learning multimodal
  representations for contact-rich tasks,'' \emph{IEEE Transactions on
  Robotics}, vol.~36, no.~3, pp. 582--596, 2020.

\bibitem{johannsmeier2019framework}
L.~Johannsmeier, M.~Gerchow, and S.~Haddadin, ``A framework for robot
  manipulation: Skill formalism, meta learning and adaptive control,'' in
  \emph{2019 International Conference on Robotics and Automation (ICRA)}.\hskip
  1em plus 0.5em minus 0.4em\relax IEEE, 2019, pp. 5844--5850.

\bibitem{wu2022primlafd}
Z.~Wu, W.~Lian, C.~Wang, M.~Li, S.~Schaal, and M.~Tomizuka, ``Prim-lafd: A
  framework to learn and adapt primitive-based skills from demonstrations for
  insertion tasks,'' 2022.

\bibitem{pmlr-v155-voigt21a}
\BIBentryALTinterwordspacing
F.~Voigt, L.~Johannsmeier, and S.~Haddadin, ``Multi-level structure vs.
  end-to-end-learning in high-performance tactile robotic manipulation,'' in
  \emph{Proceedings of the 2020 Conference on Robot Learning}, ser. Proceedings
  of Machine Learning Research, J.~Kober, F.~Ramos, and C.~Tomlin, Eds., vol.
  155.\hskip 1em plus 0.5em minus 0.4em\relax PMLR, 16--18 Nov 2021, pp.
  2306--2316. [Online]. Available:
  \url{https://proceedings.mlr.press/v155/voigt21a.html}
\BIBentrySTDinterwordspacing

\bibitem{vecerik2019practical}
M.~Vecerik, O.~Sushkov, D.~Barker, T.~Roth{\"o}rl, T.~Hester, and J.~Scholz,
  ``A practical approach to insertion with variable socket position using deep
  reinforcement learning,'' in \emph{2019 international conference on robotics
  and automation (ICRA)}.\hskip 1em plus 0.5em minus 0.4em\relax IEEE, 2019,
  pp. 754--760.

\bibitem{schoettler2020deep}
G.~Schoettler, A.~Nair, J.~Luo, S.~Bahl, J.~A. Ojea, E.~Solowjow, and
  S.~Levine, ``Deep reinforcement learning for industrial insertion tasks with
  visual inputs and natural rewards,'' in \emph{2020 IEEE/RSJ International
  Conference on Intelligent Robots and Systems (IROS)}.\hskip 1em plus 0.5em
  minus 0.4em\relax IEEE, 2020, pp. 5548--5555.

\bibitem{schoettler2020meta}
G.~Schoettler, A.~Nair, J.~A. Ojea, S.~Levine, and E.~Solowjow,
  ``Meta-reinforcement learning for robotic industrial insertion tasks,'' in
  \emph{2020 IEEE/RSJ International Conference on Intelligent Robots and
  Systems (IROS)}.\hskip 1em plus 0.5em minus 0.4em\relax IEEE, 2020, pp.
  9728--9735.

\bibitem{ma2020efficient}
Y.~Ma, D.~Xu, and F.~Qin, ``Efficient insertion control for precision assembly
  based on demonstration learning and reinforcement learning,'' \emph{IEEE
  Transactions on Industrial Informatics}, vol.~17, no.~7, pp. 4492--4502,
  2020.

\bibitem{dong2021tactile}
S.~Dong, D.~K. Jha, D.~Romeres, S.~Kim, D.~Nikovski, and A.~Rodriguez,
  ``Tactile-rl for insertion: Generalization to objects of unknown geometry,''
  in \emph{2021 IEEE International Conference on Robotics and Automation
  (ICRA)}, 2021, pp. 6437--6443.

\bibitem{gao2021kpam}
W.~Gao and R.~Tedrake, ``kpam 2.0: Feedback control for category-level robotic
  manipulation,'' \emph{IEEE Robotics and Automation Letters}, vol.~6, no.~2,
  pp. 2962--2969, 2021.

\bibitem{brahmbhatt2015occlusion}
\BIBentryALTinterwordspacing
S.~Brahmbhatt, H.~B. Amor, and H.~Christensen, ``Occlusion-aware object
  localization, segmentation and pose estimation,'' in \emph{Proceedings of the
  British Machine Vision Conference (BMVC)}, X.~Xie, M.~W. Jones, and G.~K.~L.
  Tam, Eds.\hskip 1em plus 0.5em minus 0.4em\relax BMVA Press, September 2015,
  pp. 80.1--80.13. [Online]. Available:
  \url{https://dx.doi.org/10.5244/C.29.80}
\BIBentrySTDinterwordspacing

\bibitem{grady2021contactopt}
P.~Grady, C.~Tang, C.~D. Twigg, M.~Vo, S.~Brahmbhatt, and C.~C. Kemp,
  ``{ContactOpt}: Optimizing contact to improve grasps,'' in \emph{The IEEE
  Conference on Computer Vision and Pattern Recognition (CVPR)}, 2021.

\bibitem{moveit0}
D.~Coleman, I.~Sucan, S.~Chitta, and N.~Correll, ``Reducing the barrier to
  entry of complex robotic software: a moveit! case study,'' \emph{arXiv
  preprint arXiv:1404.3785}, 2014.

\bibitem{moveit1}
I.~A. Sucan and S.~Chitta, ``{MoveIt},'' \url{https://moveit.ros.org}, 2014,
  [Online; accessed 2022-06-11].

\bibitem{sac}
\BIBentryALTinterwordspacing
T.~Haarnoja, A.~Zhou, K.~Hartikainen, G.~Tucker, S.~Ha, J.~Tan, V.~Kumar,
  H.~Zhu, A.~Gupta, P.~Abbeel, and S.~Levine, ``Soft actor-critic algorithms
  and applications,'' \emph{CoRR}, vol. abs/1812.05905, 2018. [Online].
  Available: \url{http://arxiv.org/abs/1812.05905}
\BIBentrySTDinterwordspacing

\bibitem{mnih2015human}
V.~Mnih, K.~Kavukcuoglu, D.~Silver, A.~A. Rusu, J.~Veness, M.~G. Bellemare,
  A.~Graves, M.~Riedmiller, A.~K. Fidjeland, G.~Ostrovski, \emph{et~al.},
  ``Human-level control through deep reinforcement learning,'' \emph{nature},
  vol. 518, no. 7540, pp. 529--533, 2015.

\bibitem{hausknecht2015deep}
M.~Hausknecht and P.~Stone, ``{Deep recurrent Q-learning for partially
  observable MDPs},'' in \emph{2015 {AAAI} fall symposium series}, 2015.

\bibitem{residual_rl}
T.~Johannink, S.~Bahl, A.~Nair, J.~Luo, A.~Kumar, M.~Loskyll, J.~A. Ojea,
  E.~Solowjow, and S.~Levine, ``Residual reinforcement learning for robot
  control,'' in \emph{2019 International Conference on Robotics and Automation
  (ICRA)}, 2019, pp. 6023--6029.

\bibitem{osc}
O.~Khatib, ``A unified approach for motion and force control of robot
  manipulators: The operational space formulation,'' \emph{IEEE Journal on
  Robotics and Automation}, vol.~3, no.~1, pp. 43--53, 1987.

\bibitem{florensa2017reverse}
C.~Florensa, D.~Held, M.~Wulfmeier, M.~Zhang, and P.~Abbeel, ``Reverse
  curriculum generation for reinforcement learning,'' in \emph{Conference on
  robot learning}.\hskip 1em plus 0.5em minus 0.4em\relax PMLR, 2017, pp.
  482--495.

\bibitem{sandha2021sim2real}
S.~S. Sandha, L.~Garcia, B.~Balaji, F.~Anwar, and M.~Srivastava, ``Sim2real
  transfer for deep reinforcement learning with stochastic state transition
  delays,'' in \emph{Conference on Robot Learning}.\hskip 1em plus 0.5em minus
  0.4em\relax PMLR, 2021, pp. 1066--1083.

\bibitem{andrychowicz2020learning}
O.~M. Andrychowicz, B.~Baker, M.~Chociej, R.~Jozefowicz, B.~McGrew,
  J.~Pachocki, A.~Petron, M.~Plappert, G.~Powell, A.~Ray, \emph{et~al.},
  ``Learning dexterous in-hand manipulation,'' \emph{The International Journal
  of Robotics Research}, vol.~39, no.~1, pp. 3--20, 2020.

\bibitem{schulman2017proximal}
J.~Schulman, F.~Wolski, P.~Dhariwal, A.~Radford, and O.~Klimov, ``Proximal
  policy optimization algorithms,'' \emph{arXiv preprint arXiv:1707.06347},
  2017.

\bibitem{robosuite2020}
Y.~Zhu, J.~Wong, A.~Mandlekar, and R.~Mart\'{i}n-Mart\'{i}n, ``robosuite: A
  modular simulation framework and benchmark for robot learning,'' in
  \emph{arXiv preprint arXiv:2009.12293}, 2020.

\bibitem{mujoco}
E.~Todorov, T.~Erez, and Y.~Tassa, ``Mujoco: A physics engine for model-based
  control,'' in \emph{2012 IEEE/RSJ International Conference on Intelligent
  Robots and Systems}, 2012, pp. 5026--5033.

\bibitem{TFAgents}
S.~Guadarrama, A.~Korattikara, O.~Ramirez, P.~Castro, E.~Holly, S.~Fishman,
  K.~Wang, E.~Gonina, N.~Wu, E.~Kokiopoulou, L.~Sbaiz, J.~Smith, G.~Bartók,
  J.~Berent, C.~Harris, V.~Vanhoucke, and E.~Brevdo, ``{TF-Agents}: A library
  for reinforcement learning in tensorflow,''
  \url{https://github.com/tensorflow/agents}, 2018, [Online; accessed
  2022-06-12].

\bibitem{reverb}
A.~Cassirer, G.~Barth-Maron, E.~Brevdo, S.~Ramos, T.~Boyd, T.~Sottiaux, and
  M.~Kroiss, ``Reverb: A framework for experience replay,'' 2021.

\bibitem{panda_robot}
{Franka Emika}, ``Panda research v2,'' \url{https://www.franka.de/research},
  2022, [Online; accessed 2022-06-11].

\bibitem{soft_gripper}
{Soft Robotics Inc.}, ``{mGrip} soft gripper and controller system,''
  \url{https://www.softroboticsinc.com/products/mgrip-modular-gripping-solution-for-food-automation},
  2022, [Online; accessed 2022-06-11].

\bibitem{aruco0}
F.~J. Romero-Ramirez, R.~Mu{\~n}oz-Salinas, and R.~Medina-Carnicer, ``Speeded
  up detection of squared fiducial markers,'' \emph{Image and vision
  Computing}, vol.~76, pp. 38--47, 2018.

\bibitem{aruco1}
S.~Garrido-Jurado, R.~Mu{\~n}oz-Salinas, F.~J. Madrid-Cuevas, and
  R.~Medina-Carnicer, ``Generation of fiducial marker dictionaries using mixed
  integer linear programming,'' \emph{Pattern recognition}, vol.~51, pp.
  481--491, 2016.

\bibitem{marvel2018multi}
J.~A. Marvel, R.~Bostelman, and J.~Falco, ``Multi-robot assembly strategies and
  metrics,'' \emph{ACM Computing Surveys (CSUR)}, vol.~51, no.~1, pp. 1--32,
  2018.

\end{thebibliography}

\end{document}